\documentclass{article}

\usepackage{arxiv}
\usepackage[utf8]{inputenc} 
\usepackage[T1]{fontenc}    
\usepackage{hyperref}       
\usepackage{url}            
\usepackage{booktabs}       
\usepackage{amsfonts}       
\usepackage{nicefrac}       
\usepackage{microtype}      
\usepackage{graphicx}
\usepackage{natbib}
\usepackage{doi}
\usepackage{CJKutf8}        

\title{Quality Estimation Using Round-trip Translation with Sentence Embeddings}

\date{December, 2020}	

\author{\hspace{1mm}Nathan Crone \\
    	School of Computing\\
        Dublin City University\\
    	Dublin, Ireland \\
    	\texttt{nathan.crone2@mail.dcu.ie} \\
	\And
    	\hspace{1mm}Adam Power \\
    	School of Computing\\
        Dublin City University\\
    	Dublin, Ireland \\
    	\texttt{adam.power48@mail.dcu.ie} \\
	\And
    	\hspace{1mm}John Weldon \\
    	School of Computing\\
        Dublin City University\\
    	Dublin, Ireland \\
    	\texttt{john.weldon8@dcu.ie} \\
}

\renewcommand{\shorttitle}{\textit{arXiv} Template}

\hypersetup{
    pdftitle={Quality Estimation Using Round-trip Translation with Sentence Embeddings},
    pdfsubject={cs.CL, cs.LG},
    pdfauthor={Nathan Crone, Adam Power, John Weldon},
    pdfkeywords={Quality Estimation, Machine Translation},
}

\begin{document}
\maketitle

\begin{abstract}
Estimating the quality of machine translation systems has been an ongoing challenge for researchers in this field. Many previous attempts at using round-trip translation as a measure of quality have failed, and there is much disagreement as to whether it can be a viable method of quality estimation. In this paper, we revisit round-trip translation, proposing a system which aims to solve the previous pitfalls found with the approach. Our method makes use of recent advances in language representation learning to more accurately gauge the similarity between the original and round-trip translated sentences. Experiments show that while our approach does not reach the performance of current state of the art methods, it may still be an effective approach for some language pairs.
\end{abstract}


\section{Introduction}

The purpose of a quality estimation (QE) system is to determine the quality of a translation without using a reference translation. Traditionally, translation evaluation metrics like BLEU, NIST, METEOR, and TER would be used to assess a machine translation (MT) system's translation by comparing the translation output to a reference translation created by an expert. However, quality estimation systems eliminate the need for the time-consuming task of manually annotating sentence translations to create this reference. QE “is concerned about predicting the quality of a system’s output for a given input, without any information about the expected output” \citep{Specia-etal-2010}. Quality can be measured in a number of different ways, but in this paper, we will explore the use of round-trip translation in conjunction with sentence encoding models as a means of measuring this.

Our approach is to translate an existing translated sentence back into its original language, generate sentence embedding vectors for both the original and round-trip translated sentences, and measure the similarity between the two vectors as a proxy for translation quality. Our justification behind this approach is that if we use an accurate model to translate the sentence back into the original language, the meaning of the sentence will be maintained. Then when we measure the similarity between the original sentence and the round-trip translated sentence, this score will correlate to the quality of the translation model that created the sentence pairs in the first place.

Similar to the approach used in \citet{Somers-2005}, evaluation metrics, such as BLEU, could then be used to evaluate the similarity of these two same language sentences. Nevertheless, we hypothesise that our approach of using sentence embeddings will prove far more useful than these as it can take the semantics of the sentence into consideration along with judging the sentence as a whole rather than word by word. We also think our approach will be an effective way to get around the poor prediction encountered by \citet{Somers-2005}. While alternate methods using more complex models have been used in recent years, we hope that the simplicity of our proposition and the novelty of using sentence encoding will bring new insights to this approach and will re-encourage further exploratory research in this area.

The remainder of the paper is organised as follows: In Section \ref{related} we look at existing literature, in Section \ref{method} we present our methods, Section \ref{data} details the data used, Section \ref{results} shows and discusses the results and finally, Section \ref{conclusion} concludes our research and proposes future directions for building on our research.

\section{Related Work}
\label{related}

From researching previous works, we saw from QE systems like QuEst \citep{specia-etal-2013} and QuEst++ \citep{specia-etal-2015} that machine learning algorithms, like support vector machines and decision trees, have proved quite successful as quality estimation tools in situations where a lot of work on feature extraction was already done. While techniques like this are no longer state-of-the-art, they paved a useful stepping stone for new approaches. Neural-based QE systems, such as POSTECH \citep{kim-etal-2017}, which require no feature extraction, have become more popular in research in recent years. One downfall of these methods is that they require a lot more data and take far longer to train. This need for large amounts of parallel data can cause problems which is where systems like TransQuest \citep{ranasinghe-transquest-2020} come in. This model bridges the data gap by leveraging pre-trained cross-lingual transformers that are then fine-tuned to the quality estimation task using the smaller amount of training data available. 

Early attempts at using round-trip translation (RTT) as a means of quality estimation proved unsuccessful, resulting in researchers labeling it as ‘good for nothing’ \citep{Somers-2005}. Its utility as a viable way to estimate the quality of a machine translation system has been pulled into question by problems such as those highlighted in \citet{oconnell-2001} which assert that round-trip translation relies heavily on a model accurately translating the translated sentence back to its source language. Evaluation metrics cannot tell if translation errors occurred during the first translation or during the translation back to the original language. Moreover, if an error does occur during the first translation, it may cause even more problems when it is translated back. Some researchers have attempted to address this problem by using multiple machine translation models to translate the sentences back to their original language so that errors in one translation system are reduced by the collection of models used, however, in \citet{Zaanen-2006}, they still concluded that round-trip translation was not a good way to measure machine translation quality.

One common theme we saw throughout these research papers was that prior research mainly focused on the round-trip translation and did not focus too heavily on how they were evaluating the similarity between these sentences. Evaluation metrics like BLEU are limited as they only measure the lexical similarity and do not take the semantics of the sentences into consideration. Maintaining the semantic information of a sentence is something we should want to ensure that a machine translation system does when evaluating its translations. While \citet{Somers-2005} does conclude that the results given by his RTT experiment do not look promising, he also recognises in his conclusion that there is a big reliance on BLEU and F-Score throughout his paper and he says that he would like to replicate his experiments using human rating of intelligibility. 

During the course of our work, \citet{moon-etal-2020} was published which very closely resembles our approach. They report a Pearson R correlation of 0.95 on the WMT19 metrics task evaluation set for English-German sentence pairs. We were unable to compare our approach on this data, as the ground truth direct assessment scores were not made publicly available.

\section{Our Methodology}
\label{method}

\begin{figure}[h]
    \centering
    \fbox{\includegraphics[width=16cm]{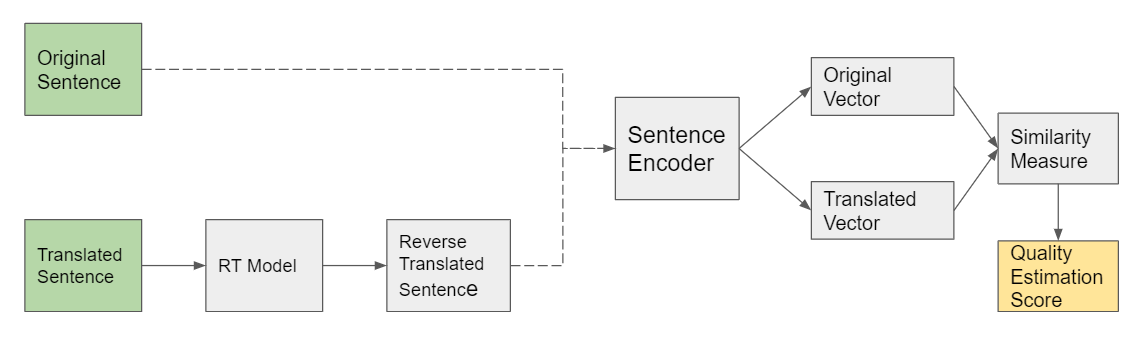}}
    \caption{Overview of our quality estimation system. Green boxes represent the input original and translated sentence pairs, the yellow box represents our final output.}
    \label{fig:MethodDiagram}
\end{figure}

\subsection {Reverse Translation (RT)}
\label{reverseTrans}

The first step in our process was to take the German sentences that the model to be estimated produced, and translate them back into English. These translations could then be used determine how similar they were to their originals. We had initially planned to build and train our own RT model, however, this was not the main aim of our research and we wanted to focus our time on experimentation with different similarity measures. As a result we decided to use the freely available and relatively simple to set up Google Translate (GT) system to achieve this based on the research by \citet{aiken-2019}. When studying the improvements google translate has made on its system over the previous 8 years, Aiken showed that that for our given language pair, German to English, GT was able to translate with an average BLEU score of 81\%. We used the associated Google Translate Python API to translate the sentences in batches.

\begin{table}[h]
    \caption{Examples of original and round-trip translated sentences.}
    \centering
    \begin{tabular}{ll}
	    \toprule
        Original English Sentence & Round-trip Translated English Sentence\\
        \midrule
        José Ortega y Gasset visited Husserl at Freiburg in 1934.& In 1934 José Ortega y Gasset visited Husserl in Freiburg.\\
        \\
        However, a disappointing ninth in China meant that he & However, a disappointing ninth in China meant he \\
        dropped back to sixth in the standings. & dropped to sixth overall. \\
        \\
        Once North Pacific salmon die off after spawning, & As soon as the North Pacific salmon dies after spawning,\\
        usually local bald eagles eat salmon carcasses almost & local bald eagles usually eat almost exclusively salmon\\
        exclusively. & bodies.\\
        \bottomrule
    \end{tabular}
    \label{tab:org_vs_roundrip}
\end{table}

\subsection {Sentence Vectorisation}
\label{sentenceVec}

We used various sentence embedding models such as Google’s Universal Sentence Encoder (USE) to vectorise our sentences \citep{cer-2018}. This was important because in order to truly compare the original English sentence to the English sentences created after our round-trip translation, we needed to have the actual meaning of the sentences to be represented in these vectors. Standard sentence vectorisation such as term frequency vectorisation essentially just produces vectors which contain information about what words are in the sentence, with no knowledge of the actual meaning of the sentence. 

If we take the following two sentences as an example:
\begin{itemize}
    \item "That phone is broken"
    \item "This iPhone is smashed"
\end{itemize}
We, as English speakers, understand immediately that these two sentences are quite similar and  would probably score this relatively high if asked how closely they match in terms of meaning. If we compared these two sentences using a similarity measure after word embedding vectorisation, we would also expect the score to be relatively high, as the meaning has been retained post vectorisation. However, because there isn't a single word that actually matches between them (once the stopwords “is”, “this” and “that” are removed), comparing the two sentences with a similarity measure after standard word vectorisation will result in a score of zero. Here is a quick explanation as to why this is.

\begin{figure}[h]
    \centering
    \fbox{\includegraphics[width=16cm]{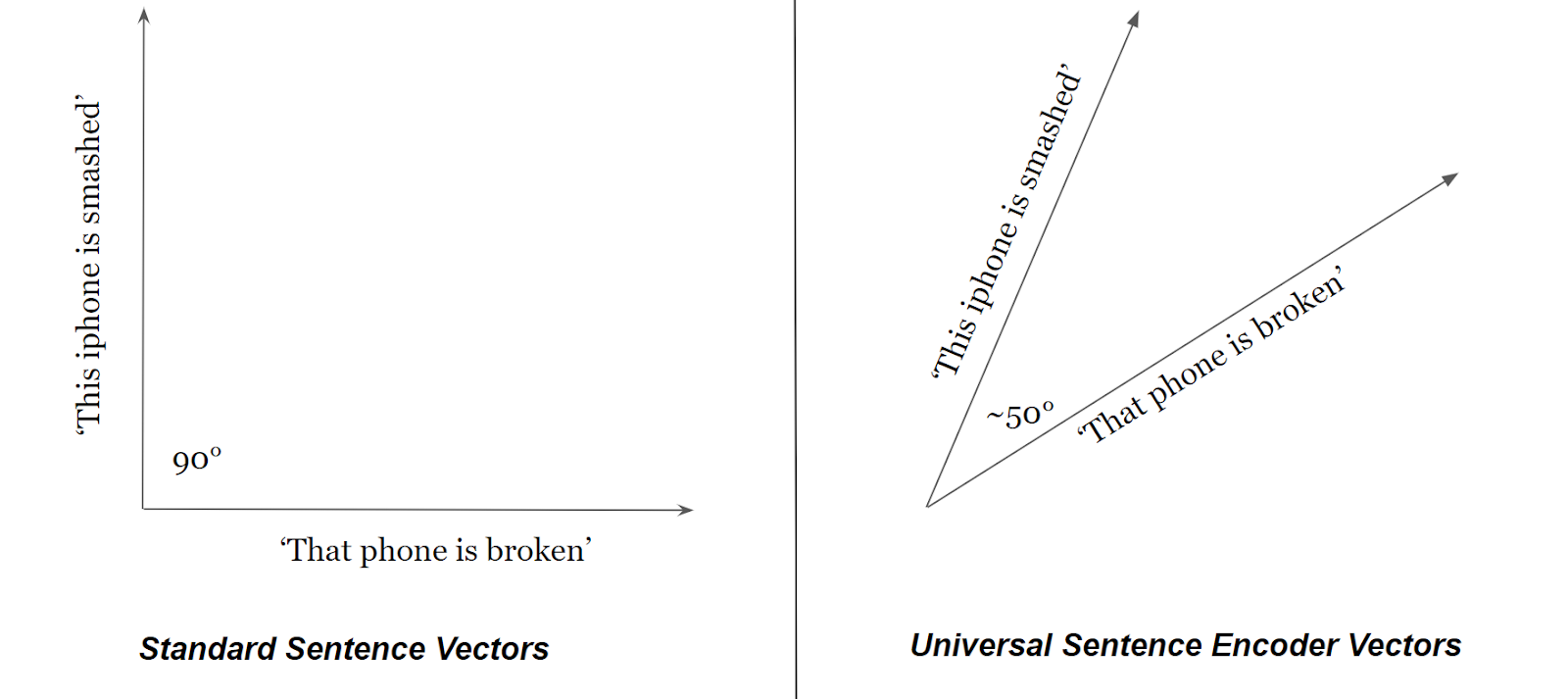}}
    \caption{Illustration of standard sentence vectors (left) and embeddings produced by USE (right).}
    \label{fig:VectorDiagram}
\end{figure}

With standard term frequency vectorisation, our two vectors will both have a length equal to the total number of terms across both sentences. In this example the full list of words after we remove stop words is 4 - [“phone”, “broken”, “iPhone”, “smashed”]. The vector representation for each sentence is then just a count, at each index, of how many times each of these terms shows up in the sentence.
\begin{itemize}
    \item “Phone broken” - [1, 1, 0, 0]
    \item “iPhone smashed” - [0, 0, 1, 1]
\end{itemize}
From this, these two vectors will be orthogonal vectors and hence will have a similarity score of zero. This standard sentence vectorisation approach is further illustrated in figure \ref{fig:VectorDiagram}. This figure also shows that when we produce two sentence vectors using USE vectorisation, these vectors have a smaller angle between them, signifying at least some similarity. Table \ref{tab:cosSimTable} helps to illustrate this more clearly by detailing the scores for this sentence comparison along with the scores for some other examples. As a result, we decided to use USE as our similarity measure for this research.

\begin{table}[h]
    \caption{Comparison of cosine similarities produced by standard sentence vectors and USE embeddings.}
    \centering
    \begin{tabular}{llll}
        \toprule
        Sentence A & Sentence B & USE Score & Standardised Cosine Score \\
        \midrule
        the boys love football & the guys love sport & 0.637 & 0.333 \\
        the phone is broken & this iphone is smashed & 0.505 & 0.000 \\
        it took too long to arrive & the delivery was late & 0.421 & 0.000 \\
        \bottomrule
    \end{tabular}
    \label{tab:cosSimTable}
\end{table}

\subsection {Sentence Vector Similarity Measure - Cosine Similarity}
\label{sentenceVecSim}

\begin{figure}[h]
    \centering
    \fbox{\includegraphics[width=16cm]{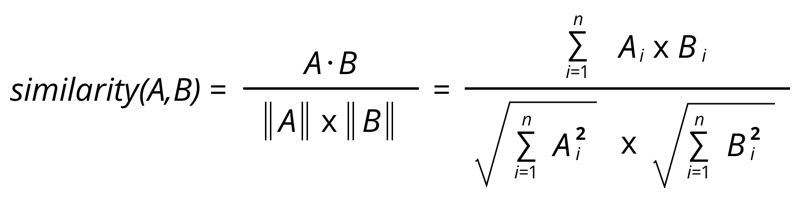}}
    \caption{Formula for the cosine similarity between two vectors A and B.}
    \label{fig:CosSimImg}
\end{figure}

As shown in figure \ref{fig:CosSimImg}, the cosine similarity between any two vectors A and B is given as the inner (dot) product of A and B, divided by the product of their Euclidean norms. The cosine of the angle between any two vectors is restricted to the range [-1,1] and this proves quite useful for similarity measures as it gives a quick, clear indication of how similar two vectors are.
\begin{itemize}
    \item 1 indicates that the two vectors are perfectly similar
    \item 0 shows that they are orthogonal vectors, having nothing in common at all
    \item -1 suggest that both vectors are totally opposed to each other
\end{itemize}
With this in mind, it should be obvious that for our use case, the closer the cosine similarity measure is to 1 for any sentence pair, the more meaning that was retained through the round-trip and thus the higher the quality of the machine translation being estimated.

\section{Data Used}
\label{data}

For this Sentence-Level Direct Assessment task that we were given we had a choice of 6 different language pairs:
\begin{itemize}
    \item English - German
    \item English - Chinese
    \item Romanian - English
    \item Estonian - English
    \item Nepalese - English
    \item Sinhala - English
\end{itemize}
Each language pair was built from Wikipedia data and varied in terms of resources with English-German and English-Chinese being high resource, Romanian-English and Estonian-English being medium resource, and Sinhala-English and Nepalese-English being low resource.

We were limited in the choice of language pair that we could use for our process because all of the sentence embedding models that we use later were trained on English corpora. This limitation meant that we only had the choice between English to German and English to Chinese. We decided that we would choose German as we had one team member who had some experience with the language and we also felt that, as native English speakers, German would be easier for us to grasp because English and German are both Germanic languages, unlike Chinese.

The data that we used throughout the process were the English to German translation sentence pairs from the “EMNLP 2020 Fifth Conference on Machine Translation (WMT20)”. This data consists of 7000 rows and 6 columns, where each row contains a translation sentence pair and various quality measures. These columns were as follows:
\begin{itemize}
    \item \textit{Original} - The original English sentence, as a string
    \item \textit{Translation} - The German translation that of the original English sentence, as a string
    \item \textit{Scores} - A list of scores for the quality of the translation, from human translators
    \item \textit{Mean} - The average of these human scores
    \item \textit{Z Scores} - These are the normalised scores given by each particular human translator
    \item \textit{Z Mean} - The average of these Z Scores
\end{itemize}
The manual scoring uses a form of direct assessment, with the process being originally laid out in the FLORES \citep{guzman-2019} setup. In this setup, professional translators evaluate each translation on a scale of 0-100. 0-10 signifies that the translation is completely inaccurate and incorrect, 70-90 represents a translation that closely resembles the original and 90-100 is seen as being a perfect translation of the original. The paper doesn’t explain what the range between 10 and 70 represents but we made the assumption that a translation is more accurate and similar to the original if the score is closer to 70 and more inaccurate and different from the original if the score is closer to 10.

\section{Experiments and Results}
\label{results}

To create the sentence embeddings used to compare the original and reverse translated English sentences, we used Google’s Universal Sentence Encoder (USE) \citep{cer-2018} as a baseline. This generated a 512 dimensional vector for each sentence which we then compared using cosine similarity. Our initial experiment showed a correlation of just 0.07 between the USE based similarity scores and the z-normalised direct assessment (DA) scores. This highlighted that our initial approach was not much better than random guessing. In order to understand why our approach did not work, we manually examined a number of the sentence pairs and their corresponding DA and similarity scores. Through this, we discovered some situations where there were clear discrepancies between the DA scores and our model scores. These broadly fell into two categories: failed forward translations and code switching in the original sentence.

\subsection{Failed Forward Translation}
\label{filed_forward_trans}

\begin{figure}[h]
    \centering
    \fbox{\includegraphics[width=16cm]{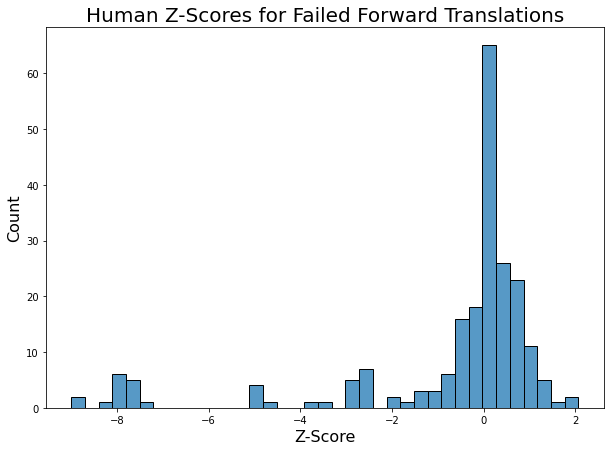}}
    \caption{Distribution of z-normalised human DA scores for failed forward translations. The “Failed” group was identified by English and German sentence pairs with BLEU scores greater than 95.}
    \label{fig:zScoreGraph}
\end{figure}

Failed forward translation refers to when the forward translation (FT) model, whose quality we are trying to estimate, has failed to translate the original sentence and instead simply copied it word for word. As the “translated” sentence is then still in the source language, our RT model does not need to translate it back again, thus leaving us with two identical sentences which of course score a perfect similarity. In our case, this situation seemed to mainly occur where the sentences consisted of names of books, people, works of art, places and institutions. Both of the arguments could be made that these names and titles should be left in their source language or that where possible, they should be translated. For example, should “University of California” and “The Wind that Shakes the Barley” be translated into “Universität von Kalifornien” and “Der Wind, der die Gerste schüttelt” or left as is? This ambiguity was clearly an issue in the manual scoring too, as the DA scores were widely distributed for these situations as shown in figure \ref{fig:zScoreGraph}. 

\subsection{Code Switching}
\label{code_switching}

\begin{CJK*}{UTF8}{gbsn}
    \begin{table}[!h]
        \caption{Example of code switching in original English sentence.}
        \centering
        \begin{tabular}{ll}
            \toprule
            Original English & Monkeys in chorus cry; Tigers and leopards roar 猿狖群嘯兮虎豹原. \\
            FT German & Monkeys in chorus schrei; Tigers and leopards brüllen, was sie bei der Suche nach bestimmten \\
            & Kriterien erwarten.\\
            RT English & Monkeys in chorus scream; Tigers and leopards roar what to expect when looking for certain \\
             & criteria.\\
            Mean DA Score & 81 \\
            Mean Z Score & 0.44 (above average) \\
            Our Z Score & -7.55 (far below average) \\
            \bottomrule
        \end{tabular}
        \label{tab:code_switching_example}
    \end{table}
\end{CJK*}

The second situation we discovered where our approach will likely fail is when there is code switching in the original sentence. Code switching is where there are multiple languages present in a sentence. There was only one example of this in the dataset, but it highlighted this as a problem if our system was applied to this kind of data. Table \ref{tab:code_switching_example} shows this example. The original sentence contains both English and Chinese words. The FT model successfully translated both the English and Chinese parts into German and the RT model then translated the fully German sentence back into English. As our system was now comparing a mixed language sentence with an English sentence, it scored their similarity very low. 

\subsection{Dataset Outliers}
\label{outliers}

There were also a lot of cases where it was not clear why our approach failed. It was very difficult to tell what exactly the manual scores represented. While there will always be some outliers in any dataset, because the manual scores were an average of at least three reviewers, their seeming abundance in this dataset was surprising to us. At times we found ourselves agreeing with the scores of our model more than those of the human experts. Examples of these cases can be seen in tables \ref{tab:sentence_example_one} and \ref{tab:sentence_example_two}.

\begin{table}[!h]
    \caption{Example showing a clear over-scoring from the human evaluators where the FT model failed to retain any semantic meaning of the original sentence.}
    \centering
    \begin{tabular}{ll}
        \toprule
        Original English & Prior to 1890, traditional religious beliefs included [[Wakan Tanka]].\\
        FT German & [[Wakan Tanka]] ist ein [[Wakan Tanka | Wakan Tanka]]. \\
        RT English & [[Wakan Tanka]] is a [[Wakan Tanka | Wakan Tanka]].\\
        Mean DA Score & 87 \\
        Mean Z-Score & 0.36 (slightly above average)\\
        Our Z-Score & -7.55 (far below average)\\
        \bottomrule
    \end{tabular}
    \label{tab:sentence_example_one}
\end{table}

\begin{table}[!h]
    \caption{Example showing a seemingly overly harsh human score, where the semantics of the sentence are maintained but there may be grammatical errors.}
    \centering
    \begin{tabular}{ll}
        \toprule
        Original English & He vacated the WWA  Cruiserweight title in April 2002 after returning to WWF.\\
        FT German & Nach seiner Rückkehr zum WWF verließ er im April 2002 den Titel WWA Cruiserweight.\\
        RT English & After returning to the WWF, he left the WWA Cruiserweight title in April 2002.\\
        Mean DA Score & 12.7 \\
        Mean Z-Score & -7.54 (far below average)\\
        Our Z-Score & 0.56 (above average)\\
        \bottomrule
    \end{tabular}
    \label{tab:sentence_example_two}
\end{table}

\subsection{Pre-trained Encoder Models}
\label{encoders}

\begin{figure}[!h]
    \centering
    \fbox{\includegraphics[width=14cm]{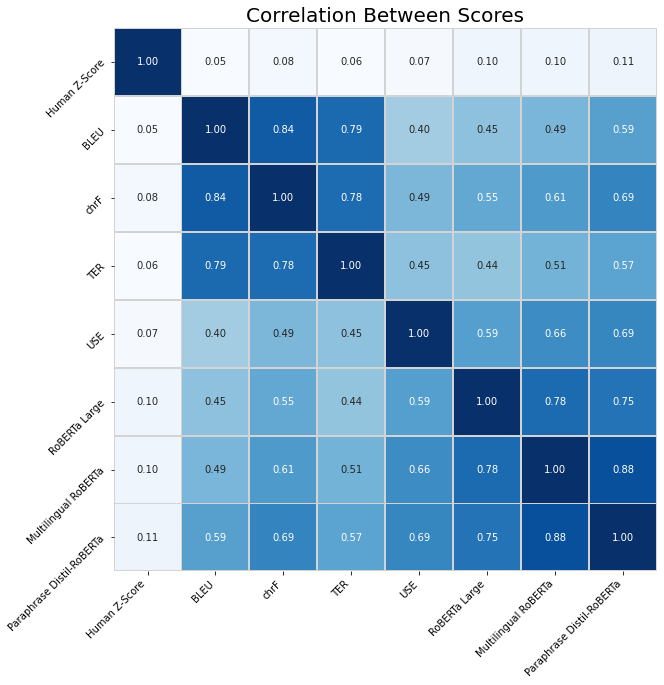}}
    \caption{Pearson R correlation between human z-normalised scores, traditional sentence similarity metrics and sentence embedding based similarity metrics.}
    \label{fig:correlationImg}
\end{figure}

To investigate the importance of the choice of sentence encoder for our approach, we tested several different pre-trained encoder models. These models were trained on different tasks, so each one may capture a different aspect of translation quality. Table \ref{tab:modelTable} outlines the details of these models. The resulting sentence embedding vectors were compared using cosine similarity as described in the methodology section. We also tested several widely used traditional sentence similarity measures, namely BLEU \citep{papineni-2002}, chrF \citep{popovic-2015}, and TER \citep{Snover-2006} scores. The results of these experiments against human scores are outlined in tables \ref{tab:resultsTable1} and \ref{tab:resultsTable2}, and in further detail in figure \ref{fig:correlationImg}. As expected, we observed the lowest correlations for the traditional measures, however, with our baseline USE based metric, we surprisingly saw a similar low correlation. Interestingly, the correlation between the traditional and USE metrics were also low, which indicates that while they both perform similarly against the human scores, they differ in how they are doing so. The other embedding based measures had a higher correlation with the human scores, which may suggest that their individual training tasks may contribute to improved sentence representation.

\begin{table}[!h]
    \caption{Details of encoder models used in our experiments.}
    \centering
    \begin{tabular}{ll}
        \toprule
        Model Name & Training Task \\
        \midrule
        Universal Sentence Encoder & Sentiment Prediction, Semantic Similarity\\
        RoBERTa Large \citep{liu-2019} & Semantic Similarity\\
        Multilingual RoBERTa \citep{conneau-2020} & Semantic Similarity (EN, DE)\\
        Paraphrase Distil-RoBERTa \citep{sanh-2020} & Paraphrase Identification\\
        \bottomrule
    \end{tabular}
    \label{tab:modelTable}
\end{table}

\begin{table}[!h]
    \caption{Pearson R correlation between human z-scores and English-English sentence similarity measures.}
    \centering
    \begin{tabular}{ll}
        \toprule
        Metric & Correlation with Human Z-Scores \\
        \midrule
        BLEU & 0.05 \\
        chrF & 0.08 \\
        TER  & 0.06 \\
        \bottomrule
    \end{tabular}
    \label{tab:resultsTable1}
\end{table}

\begin{table}[!h]
    \caption{Pearson R correlation between human z-scores and English-English sentence similarity measures.}
    \centering
    \begin{tabular}{ll}
        \toprule
        Encoder & Correlation with Human Z-Scores\\
        \midrule
        USE & 0.07 \\
        RoBERTa Large & 0.10 \\
        Multilingual RoBERTa & 0.10 \\
        Paraphrase Distil-RoBERTa & 0.11 \\
        \bottomrule
    \end{tabular}
    \label{tab:resultsTable2}
\end{table}

\subsection{Other Languages}
\label{others}

We also tested our approach on the WMT20 English-Chinese sentence level quality estimation dataset. While we did not perform in depth experiments due to time constraints, we used our baseline USE similarity measure in the same way as for the English-German dataset. We achieved a correlation of 0.25 which is higher than the score of 0.19 from the baseline neural predictor-estimator model submitted by the WMT20 organisers. This stark difference in performance of our approach across the datasets suggests that there may be a higher importance of semantic retention for the Chinese language.

\subsection{Limitations}
\label{limitations}

One of the limitations of our approach is that it relies heavily on the semantics of a sentence being retained. Grammatical mistakes made by the FT model such as punctuation, noun gender, and verb conjugation errors, may not be reflected in the translation produced by the RT model. To properly examine our approach, we would like to measure correlation on sub-categories of translation quality such as fluency, semantic retention and grammatical errors. This would give greater insight into where our approach is applicable for estimating translation quality and would highlight the characteristics of translation that it fails to capture.

\section{Conclusion}
\label{conclusion}

In this paper, we have investigated the efficacy of round-trip translation as an estimate of translation quality and proposed a novel use of sentence embeddings to measure the similarity between source and round-trip translated sentences. The aim of our approach was to use the semantic information retained through round-trip translation as a proxy for translation quality. Experimentation shows increased performance of sentence embedding based similarity measures over traditional lexical metrics. We also highlight current pitfalls of the round-trip translation approach and suggest directions for future research on this topic.

\bibliographystyle{unsrtnat}
\bibliography{qe_paper}

\end{document}